\documentclass{article}
\usepackage{spconf,amsmath,graphicx}

\usepackage{multirow}
\usepackage{graphicx} 
\usepackage{float} 
\usepackage{subfigure} 
\usepackage{booktabs}
\usepackage{colortbl}
\usepackage[export]{adjustbox}
\usepackage{caption}
\usepackage{array}
\usepackage{amsmath}
\usepackage{ragged2e}
\usepackage{amssymb}
\usepackage{adjustbox}

\title{Multi-task Transformer with Relation-attention and Type-attention for Named Entity Recognition}
\name{Ying Mo\textsuperscript{\rm 1},
    Hongyin Tang\textsuperscript{\rm 2},
    Jiahao Liu\textsuperscript{\rm 2},
    Qifan Wang\textsuperscript{\rm 3},
    Zenglin Xu\textsuperscript{\rm 4},
    \emph{Jingang Wang}\textsuperscript{\rm 2},
    \emph{Wei Wu}\textsuperscript{\rm 2},\
    \emph{Zhoujun Li}\textsuperscript{\rm 1\rm *}\thanks{ * Corresponding author}
    }
\address{
   \textsuperscript{\rm 1}{State Key Lab of Software Development Environment, Beihang University, Beijing, China}\\
   \textsuperscript{\rm 2}{Meituan, Beijing, China} \textsuperscript{\rm 3}{Meta AI, New York, United States}\\
   \textsuperscript{\rm 4}{Harbin Institute of Technology, Shenzhen, China}
    }
\begin{document}

\maketitle
\begin{abstract}
Named entity recognition (NER) is an important research problem in natural language processing. There are three types of NER tasks, including flat, nested and discontinuous entity recognition. Most previous sequential labeling models are task-specific, while recent years have witnessed the rising of generative models due to the advantage of unifying all NER tasks into the seq2seq model framework. Although achieving promising performance, our pilot studies demonstrate that existing generative models are ineffective at detecting entity boundaries and estimating entity types. This paper proposes a multi-task Transformer, which incorporates an entity boundary detection task into the named entity recognition task. More concretely, we achieve entity boundary detection by classifying the relations between tokens within the sentence. To improve the accuracy of entity-type mapping during decoding, we adopt an external knowledge base to calculate the prior entity-type distributions and then incorporate the information into the model via the self and cross-attention mechanisms. We perform experiments on an extensive set of NER benchmarks, including two flat, three nested, and three discontinuous NER datasets. Experimental results show that our approach considerably improves the generative NER model's performance.
\end{abstract}
\section{Introduction}
\label{sec:intro}
Named entity recognition (NER) is a fundamental research problem in natural language processing, which has been widely adopted in information retrieval and question answering systems ~\cite{molla2006-nerqa,li2014-incrementaljoint,chen2021-IE-frustratingly,li2020b}.
The NER task can be categorized into three categories, including flat NER, nested NER and discontinuous NER.
Previous works ~\cite{li2020b,Lample2016,strubell2017fast,2018bidirectional,wang-lu-2019} address NER tasks with the task-specific token-level sequential labeling or span-level classification methods.
In token-level sequential labeling methods, each token is assigned a label to represent its entity type.
On the other hand, span-level classification methods enumerate all possible spans in the sentences and classify them into pre-defined entity types.
One main drawback of these methods is that they are not able to tackle all three NER tasks concurrently, but build separate models for different categories.

Recently, sequence-to-sequence (seq2seq) generative approaches ~\cite{yan2021unified,zhang-debias2022} have been proposed to jointly model all NER tasks in a unified framework, which attracts a lot of attention in the NER community. 
Although achieving promising results on all three NER categories, there are two major limitations of these generative models. 
First, they are not effective at detecting entity boundaries. Generative models achieve the entity recognition task through an auto-regressive decoding process, where the token-relations (e.g., whether two tokens belong to a same entity or not) are not captured in seq2seq models. 
Second, entity-type relations are not explicitly considered in the seq2seq framework. The entity type generation is based on its compounding tokens and thus could be misguided if the compounding tokens are not generated correctly from the decoder.
Therefore, it is a critical problem to incorporate the entity boundaries and entity type mapping information into the seq2seq NER models.

In this paper, we propose a multi-task Transformer with relation attention and type attention, which introduces an additional entity boundary detection task into the seq2seq NER model.
More concretely, we achieve the purpose of entity boundary recognition by detecting the head, middle and tail tokens of entities.
To enhance the mapping capability between the entities and their corresponding types, we further incorporate entity-type information by leveraging an external knowledge base (i.e., Wikipedia).
The entity-type distributions could be integrated into the encoder and decoder conveniently with self-attention and cross-attention mechanisms.
Our contributions can be summarized as follows: (1) We propose a multi-task Transformer model for seq2seq NER, which incorporates the entity boundary recognition task into the named entity recognition task, and can unify the three different NER categories.
(2) We introduce two novel attention mechanisms, including entity token relation attention and entity type attention, which improve the performance of the seq2seq NER model.
(3) We conduct extensive experiments to demonstrate our model's effective capabilities compared with several state-of-the-art baselines.
\section{Methodology}
\begin{figure}[t] 
	\centering 
	\includegraphics[width=0.9\columnwidth]{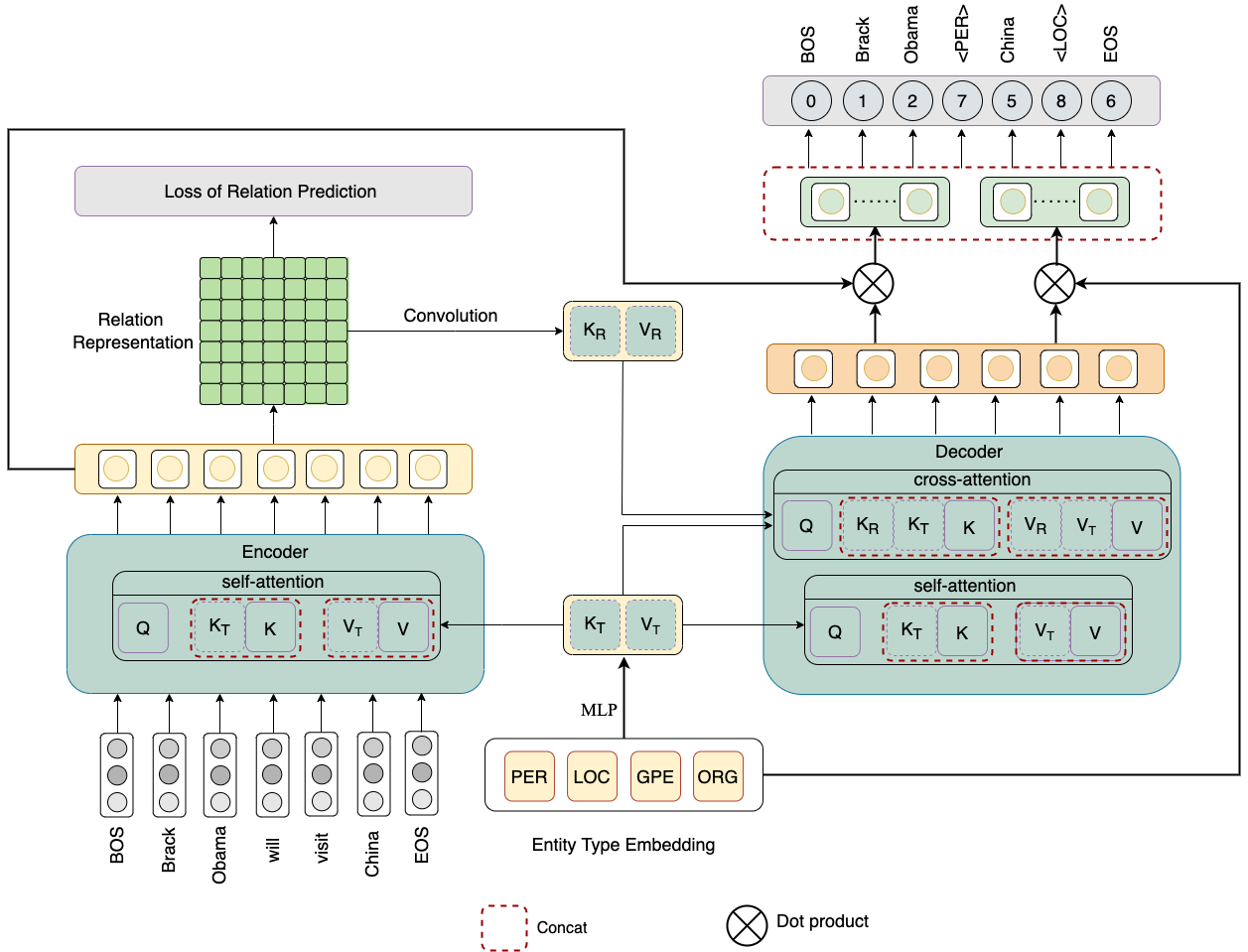} 
	\caption{The architecture of our multi-task Transformer model, which contains relation prediction and entity decoding tasks. We incorporate the entity token relation attention into the cross-attention layer of the decoder. The entity type attention is adopted to both the encoder and decoder.} 
	\label{Fig.frame} 
\end{figure}
\subsection{Approach Overview}
Figure \ref{Fig.frame} shows the overall model architecture.
Our model consists of an encoder that encodes the input sequence to its contextual embedding and a decoder that generates the output sequence with entity annotations. 
In addition to the entity generation task, we design a relation prediction task over the entity token pairs to better capture the correlations among the entities. 
The entity token relation attention and entity type attention are introduced and fused into the encoder and decoder.
We present the detail of each component separately in the following subsections.
\subsection{Relation Prediction Task}
The relation prediction task aims to learn better entity token representations for improving entity boundary detection.
Specifically, the relationship between a pair of tokens can be divided into three categories:

\noindent $\bullet$ \textbf{Begin-End}, which means that the token pair belongs to the same entity. 
One is the begin token, the other is the end token.

\noindent $\bullet$ \textbf{*-Inside}, which represents that the token pair belongs to one entity, and the two tokens are the * and inside of the entity respectively (* means begin, inside or end). 

\noindent $\bullet$ \textbf{None}, the token pair does not have any correlation.

\noindent For example, ``aching in shoulders'' is a discontinuous entity in ``I am having aching in the legs and shoulders''.
We consider the token pair (``aching'', ``shoulders'') to be the begin and end of one entity, i.e., Begin-End. 
The token pairs (``aching'', ``in'') and (``in'', ``shoulders'') are of the *-Inside relation. 
These relations among the entity tokens are useful to enable the model to understand the latent information of entity boundaries in entity token pairs.

We adopt the CLN mechanism \cite{Yu2021} to model the relations between token pairs, which could be seen as $\{r_{ij}|(i,j\in[1,N])\} \in \Bbb{R}^{N\times N\times d_h}$, where $r_{ij}$ denotes the relation representation of the token pair $(x_i,x_j)$. 
Specifically, a conditional vector is introduced as the extra contextual information to generate the gain parameter $\gamma$ and bias $\lambda$ of the layer normalization mechanism as follows:
\begin{equation}
\begin{aligned}
r_{ij} &= CLN(h_i,h_j) &= \gamma_{i} \odot (\frac{h_j-\mu }{\sigma}) + \lambda_{i}
\end{aligned}
\end{equation}
\begin{equation}
\begin{aligned}
\gamma _{ij} = W_\alpha h_i + b_\alpha, \lambda_{ij} = W_\beta h_i + b_\beta
\end{aligned}
\end{equation}
\begin{equation}
\begin{aligned}
\mu =\frac{1}{d_h}\sum_{k=1}^{d_h}h_{jk},
\sigma = \sqrt{\frac{1}{d_h}\sum_{k=1}^{d_h}(h_{jk}-\mu)^2},
\end{aligned}
\end{equation}
where $h_i$ is represetation of the token $x_i$ by the encoder. $\mu$ and $\sigma$ are the mean and standard devilation taken across the elements of $h_{j}$ respectively. $h_{jk}$ means the k-th dimension vector of $h_{j}$ and $N$ is the sequence length.

The prediction of token pair relation is calculated by MLP \cite{Liu_MLP}, And we use focal loss in order to alleviate the unbalanced distribution of token pair relations. Formulas as follows:
\begin{equation}
\begin{aligned}
p_{r_{ij}} = Softmax(MLP(r_{ij}))
\end{aligned}
\end{equation}
\begin{equation}
\begin{aligned}
L_{relation}=-\alpha(1-p_{r_{ij}})^{\tau}log(p_{r_{ij}}),
\end{aligned}
\end{equation}
where $\alpha,\tau $ are the hyperparameters. 
\subsection{Token Relation Attention}
%
With the guidance of the relation prediction task, the boundary information is embedded into the entity relation representations $r_{ij}$. 
These representations are valuable knowledge for the model to perceive and learn more accurate entity boundary,  which plays an important role in identifying entities.
Inspired by HIT \cite{WangYu2020}, we incorporate the relationship of the token pair into cross-attention layers in the decoder. 
This way, the pre-training model's structure is preserved, and entity recognition can be enhanced by comprehensively leveraging the explicit entity relation information. 

Given the relation representations $r_{ij}$, we extract and transform the feature to a vector form using a Convolutional Neural Network \cite{chiu2016} to filter out useless information for entity boundary recognition. 
Finally, the obtained entity token relation key $K_{R}$ and value $V_{R}$ matrices are concatenated with the original key $K$ and value $V$ respectively in the cross-attention layers, to enhance the model generation on entity boundaries. 
The entity relation attention is defined as:
\begin{equation}
\begin{aligned}
R_{ij}^*=Convolution(r_{ij})
\end{aligned}
\end{equation}
\begin{equation}
\begin{aligned}
head^l&=Attn(Q^l,cat(K_{R}^l,K_{T}^l,K^l), cat(V_{R}^l,V_{T}^l,V^l)),
\end{aligned}
\end{equation}
where $K_{R}^l$, $V_{R}^l,$ are from $R_{ij}^*$, $head^l$ is the $l-$th layer's head vector. $K_{T}^l,V_{T}^l,$ are calculated from the entity type embeddings described in the following section.

\subsection{Entity Type Attention}
Heterogeneous factors such as entity boundaries and types \cite{Yu2020,fu2021nested,aly2021type} have impact on entity recognition.
In this section, we discuss the modeling of entity types in the our seq2seq NER model, allowing interactions with the inputs and guiding the model to learn more effective token representation and entity recognition. These entity type representations are incorporated into both the encoder and decoder of our model through entity type attention. 

The representation of an entity type $t$ is the weighted sum of entity representations obtained from an external entity base. 
For example, assuming the entity type set $T=\{ person,location,organization,other \}$, entity type $t$ = $location$ contains entities $C_{t} = \{Beijing, Athens, London, \\ New York\}$, the initial entity type representation could be obtained as follows:
\begin{equation}
\begin{aligned}
E_t=\theta_1 E_{C_t^{1}} + ......+ \theta_iE_{C_t^{i}},
\end{aligned}
\label{fun.ET}
\end{equation}
where $E_{C_t^{i}}$ is the embedding of $C_t^{i}$ ,and $C_t^{i}$ is $i-$th entity in $C_{t}$, $\theta_i$ is the entity's weight computed from its frequency.

To leverage the entity type information in our model, we design a entity type attention in both the encoder and decoder and integrate it into the all corresponding self-attention layers. 
Two sets of the vectors $K_{T}$ and $V_{T}$ are introduced to represent the key and value of the entity type, which are concatenated with the original key $K$ and value $V$ vectors as follows:
\begin{equation}
\begin{aligned}
head^l &=Attn(Q^l,cat(K_{T}^l,K^l),cat(V_{T}^l,V^l)),
\end{aligned}
\end{equation}
where $head^l$ is the $l-$th layer's head vector. $K_{T},V_{T}$ are obtained from the embedings $E_T$, and $E_T$ is obtained by an MLP layer after formula \ref{fun.ET}. 
$K_{T}$, $V_{T}$ are split into $N_h$ head vectors respectively $K_{T}^l$, $V_{T}^l$. 
\subsection{Entity Generation}


The decoder decodes the token embeddings from the encoder to generate the entities. 
At each step $t$, the decoder obtains the token embedding $h^d_t$ based on the encoder output. Finally, the output token index distribution $P_t$ is obtained by the dot product of $h_t^d$ between $H^e$ and $E_T$ respectively.
All the previous decoded tokens $h^d_t$ and $P_t$ as:
\begin{equation}
\begin{aligned}
h_t^d =Decoder(H^e,Y_{<t}^\wedge)
\end{aligned}
\end{equation}
\begin{equation}
\begin{aligned}
P_t=Softmax(\left [ H^{e}\otimes h_t^d;E_{T}\otimes h_t^d\right ] ),
\end{aligned}
\end{equation}
where $Y_{<t}^\wedge =\left [ y_1^\wedge ,...,y_{t-1}^\wedge  \right ]$ is the generated indexes before $t$. $E_{T}$ is the embedding of the entity types. The embeddings are shared between the encoder and decoder. $\otimes$ denotes the dot product and $[.;.]$ means concatenation.

We learn to minimize the negative log-likelihood with respect to the corresponding ground-truth labels, and combine the entity generation loss and the above entity token relation loss as our final loss. The formulas are as follows:
\begin{equation}
\begin{aligned}
L_{entity}= -\log{p(Y|X)}
\end{aligned}
\end{equation}
\begin{equation}
\begin{aligned}
L = L_{entity}+ wL_{relation},
\end{aligned}
\end{equation}
where $w$ is the hyperparameter to balance the two terms.

\section{Experiments}
\subsection{Datasets}
We evaluate \textbf{flat NER datasets} on the CoNLL-2003 following the same settings in previous works \cite{Lample2016,Yu2020} and OntoNotes 5.0 which is adopted the splits as used in \cite{yan2021unified,Yu2020}.
And we experiment on \textbf{nested NER datasets} GENIA following \cite{yan2021unified,WangYu2020,zhang-debias2022}, ACE 2004, ACE 2005 in english following \cite{yan2021unified,Yu2020}. We conduct experiment on \textbf{discontinuous NER datasets} CADEC, ShARe13 and ShARe14 corpus and use \cite{Dai2020-effective} to process the data.
\subsection{Implementation Details}
We adopt BART-Large as the backbone network for all experiments.
We use AdamW \cite{Adaw2017} optimizer and the learning rate is $1e^{-5}$ for the BART-Large model and $5e^{-5}$ for other components. 
Batch size is 32 for Ontonotes 5.0 and 16 for the other datasets. 
The hyperparameters $w$ is within this range $[0.1,0.9]$. 
When the key $K_{T}$ and value $V_{T}$ of self-attention is obtained by entity type embedding, we use a MLP similar to the proj\_down-proj\_up architecture \cite{prefixtuning} with down dim 512 and up dim 1024. 
After obtaining the entity token relation representation, we reduce the dimension to 256 and use a convolution layer to convert it to the key and value required by the attention mechanism. 
$\alpha$ and $\tau$ in the focal loss of the entity relation are set to 5 and 1 respectively. 
Following the prior work \cite{yan2021unified,zhang-debias2022}, we compute the precision (P), recall (R) and F1 (F) scores and tune model according to the F1 on the development set. 
\begin{table}[ht]
\arrayrulecolor{black}
\begin{adjustbox}{width=0.85\linewidth,center}
\begin{tabular}{lllllll} 
\toprule
\multirow{2}{*}{Model}  & \multicolumn{3}{c}{CoNLL2003} & \multicolumn{3}{c}{OntoNotes 5.0}            \\
\cmidrule(lr){2-4} \cmidrule(lr){5-7}
                        &\multicolumn{1}{c}{P}     & \multicolumn{1}{c}{R}      & \multicolumn{1}{c}{F}            
                        & \multicolumn{1}{c}{P}    & \multicolumn{1}{c}{R}      & \multicolumn{1}{c}{F}      \\ 
\hline
Lample et al. (2016) \cite{Lample2016}      & \multicolumn{1}{c}{-}    & \multicolumn{1}{c}{-}     & 90.94        
                        & \multicolumn{1}{c}{-}    & \multicolumn{1}{c}{-}     & \multicolumn{1}{c}{-}      \\
Strubell et al. (2017) \cite{strubell2017fast} & \multicolumn{1}{c}{-}    & \multicolumn{1}{c}{-}     & 90.65         
                         & \multicolumn{1}{c}{-}    & \multicolumn{1}{c}{-}     & 86.84  \\
Strakova et al. (2019) \cite{strakova2019}    & \multicolumn{1}{c}{-}    & \multicolumn{1}{c}{-}     & 92.98         
                        & \multicolumn{1}{c}{-}    & \multicolumn{1}{c}{-}     & \multicolumn{1}{c}{-}      \\
Yu et al. (2021) \cite{Yu2021}          & \underline{92.91}     & 92.13    & 92.52         
                        & \underline{90.01}     & 89.77    & 89.89  \\ 
Yan et al. (2021) \cite{yan2021unified}  & 92.61   & \textbf{93.87}  & \underline{93.24}        
                        & 89.99   & 90.77           & 90.38\\
Zhang et al. (2022) \cite{zhang-debias2022}& 92.78   & 93.51           & 93.14   
                        & 89.77   & \underline{91.07}   & \underline{90.42}   \\ 
\arrayrulecolor{black}\hline
Ours                    &\textbf{93.11}  &\underline{93.77}  &\textbf{93.44} 
                        &\textbf{90.18}  &\textbf{91.08}  &\textbf{90.63}   \\
\bottomrule
\end{tabular}
\end{adjustbox}
\caption{Results for Flat NER datasets.}
\label{t1}
\end{table}
\begin{table}[ht]
\arrayrulecolor{black}
\begin{adjustbox}{width=1.0\linewidth,center}
\begin{tabular}{llllllllll} 
\toprule
\multirow{2}{*}{Model} & \multicolumn{3}{c}{ACE2004}   & \multicolumn{3}{c}{ACE2005}  & \multicolumn{3}{c}{GENIA} \\ 
\cmidrule(lr){2-4}\cmidrule(lr){5-7}\cmidrule(lr){8-10}
                       & \multicolumn{1}{c}{P}       & \multicolumn{1}{c}{R}      & \multicolumn{1}{c}{F}       
                       & \multicolumn{1}{c}{P}       & \multicolumn{1}{c}{R}      & \multicolumn{1}{c}{F}     
                       & \multicolumn{1}{c}{P}       & \multicolumn{1}{c}{R}      & \multicolumn{1}{c}{F}       \\ 
\hline
Ju et al. (2018)\cite{ju2018neural}   & \multicolumn{1}{c}{-}    & \multicolumn{1}{c}{-}   & \multicolumn{1}{c}{-} 
                       & 74.20  & 70.30  & 72.20   
                       & 78.50  & 71.30  & 74.70   \\
Strakova et al. (2019) \cite{strakova2019}   & \multicolumn{1}{c}{-}   & \multicolumn{1}{c}{-}     & 84.33  
                       & \multicolumn{1}{c}{-}   & \multicolumn{1}{c}{-}     & 83.42   
                       & \multicolumn{1}{c}{-}   & \multicolumn{1}{c}{-}     & 78.20   
                       \\
Wang et al. (2020b) \cite{WangYu2020}     & 86.08  & \underline{86.48}  & 86.28  
                       & 83.95  & 85.39  & 84.66   
                       & 79.45  & \underline{78.94}  & \underline{79.19}   
                       \\
Yu et al. (2021) \cite{Yu2021}         & 85.42  & 85.92  & 85.67  
                       & \underline{84.50}  & 84.72  & 84.61   
                       & 79.43  & 78.32  & 78.87   
                       \\ 
Yan et al. (2021) \cite{yan2021unified} & \textbf{87.27}  & 86.41  & \underline{86.84}  
                       & 83.16  & 86.38  & 84.74   
                       & 78.57  & \textbf{79.30}  & 78.93
                       \\
Zhang et al. (2022) \cite{zhang-debias2022} & 86.36  & 84.54  & 85.44   
                         & 82.92  & \underline{87.05}  & \underline{84.93}   
                         & \underline{81.04}  & 77.21  & 79.08  
                         \\ 
\arrayrulecolor{black}\hline
Ours                   & \underline{86.80}           & \textbf{87.94}  & \textbf{87.36}
                       &\textbf{84.85}   & \textbf{87.46}   & \textbf{86.14}   
                       &\textbf{81.27}   &78.33            & \textbf{79.77}  
                       \\
\bottomrule
\end{tabular}
\end{adjustbox}
\caption{Results For Nested NER datasets.}
\vspace{-2mm}
\label{t2}
\end{table}
\begin{table}[ht]
\arrayrulecolor{black}
\begin{adjustbox}{width=1.0\linewidth,center}
\begin{tabular}{llllllllll} 
\toprule
\multirow{2}{*}{Model} & \multicolumn{3}{c}{CADEC}   &\multicolumn{3}{c}{ShARe13}  & \multicolumn{3}{c}{ShARe14}  \\ 
\cmidrule(lr){2-4}\cmidrule(lr){5-7}\cmidrule(lr){8-10}
                       & \multicolumn{1}{c}{P}       & \multicolumn{1}{c}{R}      & \multicolumn{1}{c}{F}       
                       & \multicolumn{1}{c}{P}       & \multicolumn{1}{c}{R}      & \multicolumn{1}{c}{F}     
                       & \multicolumn{1}{c}{P}       & \multicolumn{1}{c}{R}      & \multicolumn{1}{c}{F}       \\ 
\hline
Metke-Jimenez et al.  \cite{metke2016concept}      & 64.40  & 56.50  & 60.20      
                       & \multicolumn{1}{c}{-}  & \multicolumn{1}{c}{-}  & \multicolumn{1}{c}{-}                  
                       & \multicolumn{1}{c}{-}  & \multicolumn{1}{c}{-}  & \multicolumn{1}{c}{-}\\
Tang et al. (2018) \cite{tang2018}       & 67.80  & 64.99  & 66.36      
                       & \multicolumn{1}{c}{-}  & \multicolumn{1}{c}{-}  & \multicolumn{1}{c}{-}                  
                       & \multicolumn{1}{c}{-}  & \multicolumn{1}{c}{-}  & \multicolumn{1}{c}{-} \\
Wang and Lu (2019) \cite{wang-lu-2019}   & \textbf{72.10}  & 48.40  & 58.00      
                       & \underline{83.80}  & 60.40  & 70.30      
                       & \textbf{79.10}  & 70.70  & 74.70    \\ 
Dai et al. (2020) \cite{Dai2020-effective} & 68.90  & 69.00  & 69.00 
                          & 80.50  & 75.00  & 77.70      
                          & 78.10   & 81.20  & 79.60       \\
Wang et al. (2021) \cite{wang2021clique} & 70.50  & \textbf{72.50}  & 71.50      
                       & \textbf{84.30}  & \underline{78.20}  & \textbf{81.20}      
                       & 78.20 & \textbf{84.70} & \underline{81.30}    \\
Yan et al. (2021) \cite{yan2021unified} & 70.08  & 71.21  & 70.64      
                       & 82.09  & 77.42  & 79.69      
                       & 77.20  & 83.75  & 80.34     \\
Zhang et al. (2022) \cite{zhang-debias2022}   & 71.35  & 71.86  & \underline{71.60}  
                           & 81.09  & 78.13  &79.58       
                           & 77.88  & 83.77  & 80.72    \\   
\arrayrulecolor{black}\hline
Ours                   & \underline{71.99}          &\underline{71.92}     &\textbf{71.96}
                       & 82.09          & \textbf{78.38}    & \underline{80.19}
                       & \underline{78.96}          & \underline{83.84}    & \textbf{81.33}         \\
\bottomrule
\end{tabular}
\end{adjustbox}
\caption{Results for Discontinuous NER datasets.}
\label{t3}
\end{table}
\subsection{Main Results}
We compare our model with three seq2seq models \cite{strakova2019,yan2021unified,zhang-debias2022}, and several other baselines that are specifically designed for individual NER substask, including sequence labeling \cite{ju2018neural}, span-based methods \cite{WangYu2020} and hypergraph model \cite{wang-lu-2019}, etc. 
The performance comparisons on three type datasets are reported in Table \ref{t1}, \ref{t2} and \ref{t3} respectively. 
There are several key observations from the comparison results. 
First, it can be seen that our model achieves the best performance among all compared methods on almost all datasets. 
E.g., on GENIA, our model outperforms the sequence labeling \cite{ju2018neural} and span-based methods \cite{WangYu2020} by $5.07\%$ and $4.67\%$ in terms of F1 score, and increases over $1.57\%$, $0.54\%$ and $0.69\%$ compared with the seq2seq models \cite{strakova2019,yan2021unified,zhang-debias2022}. 
The hypothesis is that our approach effectively models the entity relation, improving entity boundary detection for better entity generation. 
Moreover, the entity type information incorporated in our model further boosted the model performance. 
Second, we find that the improvement of our model over the baselines on discontinuous NER datasets is marginal, especially compared to \cite{wang2021clique}. 
The reason is that \cite{wang2021clique} is designed explicitly for discontinuous NER with complex multi-stage modeling, which is not generally applicable for nested NER. 
Nevertheless, our model can still achieve similar or even better performance. 
\subsection{Ablation Study}
To understand the effectiveness of different components in our model, we conduct a set of ablation studies by removing each component individually, i.e., entity token relation attention (TRA) and entity type attention (ETA).
Specifically, we take the seq2seq model without relation prediction task (RP), entity type attention (ETA), and entity token relation attention (TRA) as ``Baseline''. 
The results on different types of NER tasks are shown in Table \ref{t4.alation}. 
From these results, it is clear that the multi-task model with either entity relation attention (+ RP \& TRA) or entity type attention (+ RP \& ETA) substantially improves over the baseline model, which validates the effectiveness of both the token relation attention and the entity type attention.
Nevertheless, combining the multi-task with both components achieves the best performance.
\begin{table}[t]
\arrayrulecolor{black}
\begin{adjustbox}{width=0.75\columnwidth,center}
\begin{tabular}{llll} 
\toprule
\multirow{1}{*}{Model} & \multicolumn{1}{c}{CoNLL2003} & \multicolumn{1}{c}{ACE2004}  & \multicolumn{1}{c}{CADEC} \\ 
\hline
Baseline      
& \multicolumn{1}{c}{92.82}
& \multicolumn{1}{c}{86.43}   
& \multicolumn{1}{c}{70.36}      \\
+ RP \& TRA    
& \multicolumn{1}{c}{93.01\small{\textbf{\textcolor{blue}{$\uparrow$0.19}}}}
& \multicolumn{1}{c}{86.87\small{\textbf{\textcolor{blue}{$\uparrow$0.44}}}}   
& \multicolumn{1}{c}{71.10\small{\textbf{\textcolor{blue}{$\uparrow$0.74}}}}     \\
+ RP \& ETA        
& \multicolumn{1}{c}{93.35\small{\textbf{\textcolor{blue}{$\uparrow$0.53}}}}
& \multicolumn{1}{c}{87.11\small{\textbf{\textcolor{blue}{$\uparrow$0.68}}}}   
& \multicolumn{1}{c}{70.59\small{\textbf{\textcolor{blue}{$\uparrow$0.23}}}}     \\
+ RP \& TRA \& ETA  
&\multicolumn{1}{c}{93.44\small{\textbf{\textcolor{blue}{$\uparrow$0.62}}}}
& \multicolumn{1}{c}{87.36\small{\textbf{\textcolor{blue}{$\uparrow$0.93}}}}  
& \multicolumn{1}{c}{71.96\small{\textbf{\textcolor{blue}{$\uparrow$1.60}}}} \\
\bottomrule
\end{tabular}
\end{adjustbox}
\caption{F1 of three NER datsets based on different methods}
\vspace{-3mm}
\label{t4.alation}
\end{table}

\subsection{Impact of Multi-task Learning}
\begin{figure}[h] 
	\centering 
	\includegraphics[width=1\columnwidth]{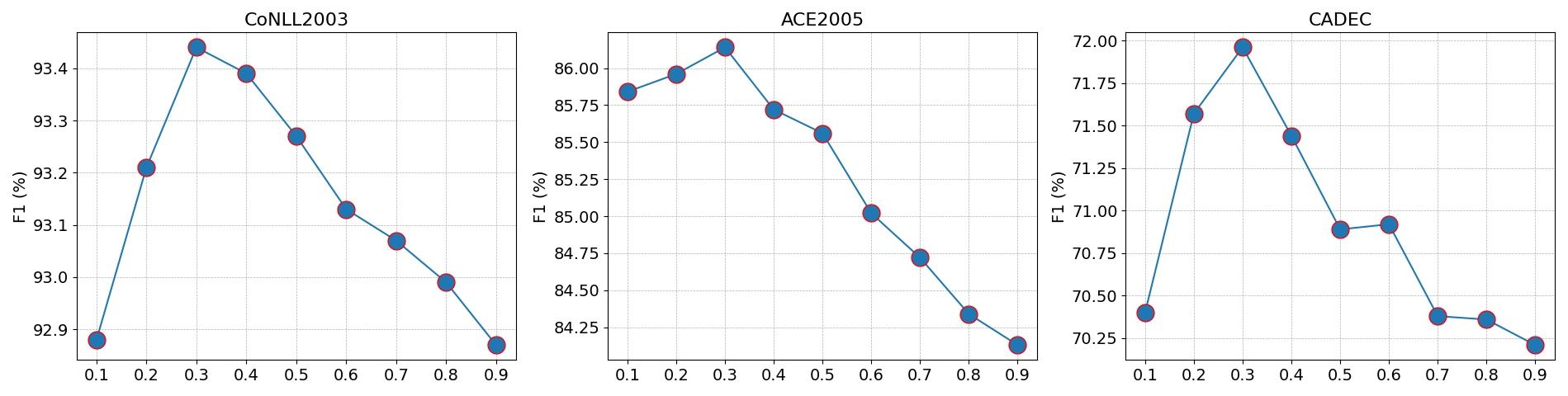} 
	\caption{The impact of multi-task learning.}
	\label{Fig.w_ablation}
\end{figure}

In order to analyze the impact of the multi-task learning, we conduct a set of experiments on CoNLL2003, ACE2005 and CADEC datasets. 
In each experiment, we modify the weight parameter $w$ from \{0.1, 0.2, 0.3, 0.4, 0.5, 0.6, 0.7, 0.8, 0.9\}, while fixing the other hyperparameters to the values as described in our implementation details. Essentially, the value of $w$ controls the importance of the two tasks.
The model performances with different task weights are shown in Figure \ref{Fig.w_ablation}. 
We find that our model achieves the best performance when $w$ is around 0.3 on all datasets. It can also been seen that the model performance drops with the increasing of $w$. Our hypothesis is that the entity generation task plays a more important role compared to the entity relation prediction task.
Therefore, it is crucial to identify a good balance between the two tasks.
\vspace{-3mm}
\section{Conclusion}
In this paper, we propose a multi-mask Transformer model that integrates the relation prediction and entity recognition task into a unified seq2seq framework. 
We further introduce two attention mechanisms, entity token relation attention and entity type attention, into the encoder and decoder to improve the performance.
Experimental results on three NER tasks demonstrate the superior performance of our model compared with several methods. 
There are several possible research directions. We plan to continue to study the seq2seq NER with large-scale pre-trained models. We also plan to conduct seq2seq NER in few-shot or zero-shot scenarios.





\vfill\pagebreak



\bibliographystyle{IEEEbib}
\bibliography{icassp23}

\end{document}